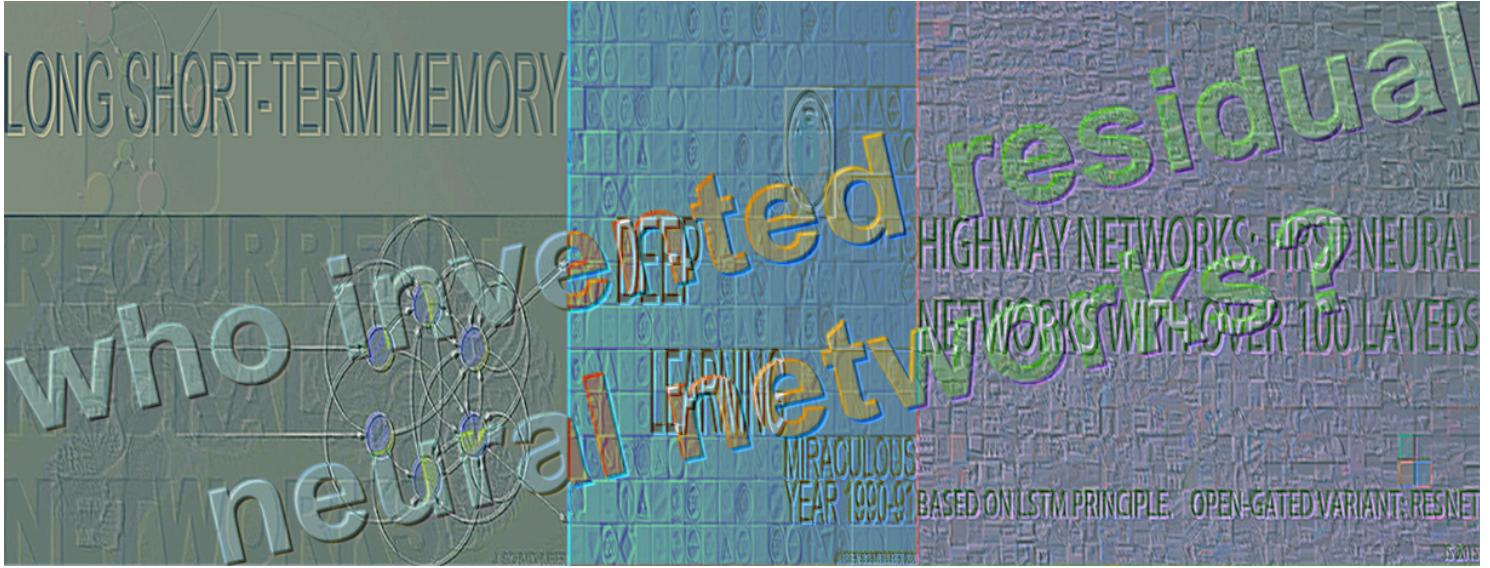

Jürgen Schmidhuber (28 Sep 2025)                              AI Blog
Pronounce: You_again Shmidhoobuh                      @SchmidhuberAI
Technical Report IDSIA-09-25, IDSIA                    juergen@idsia.ch

# Who invented deep residual learning?

Modern AI is based on deep artificial neural networks (NNs).[DLH] As of 2025, the most cited scientific article of the 21st century is an NN paper on *deep residual learning* with *residual connections*.[MOST25,25b] Who invented this? Here is the timeline of the evolution of deep residual learning:

★ 1991: recurrent residual connections (weight 1.0) solve the vanishing gradient problem
★ 1997 LSTM: *plain* recurrent residual connections (weight 1.0)
★ 1999 LSTM: *gated* recurrent residual connections (gates initially open: 1.0)
★ 2005: unfolding LSTM—from *recurrent* to *feedforward* residual NNs
★ May 2015: very deep Highway Net—gated *feedforward* residual connections (initially 1.0)
★ Dec 2015: ResNet—like an open-gated Highway Net (or an unfolded 1997 LSTM)

## 1991: recurrent residual connections solve the vanishing gradient problem

Sepp Hochreiter introduced *residual connections* for recurrent NNs (RNNs) in a diploma thesis (June 1991)[VAN1] supervised by Jürgen Schmidhuber, at a time when compute was about 10 million times more expensive than today (2025). His *recurrent residual connection* was mathematically derived from first principles to overcome the fundamental deep learning problem of vanishing or exploding gradients, first identified and analyzed in the very same thesis.[VAN1][DLP][DLH]

Like most good things, the *recurrent residual connection* is very simple: a neural unit with the *identity activation function* has a connection to itself, and the weight of this connection is 1.0.

That is, at every time step of information processing, this unit just adds its current input to its previous activation value. So it's just an incremental integrator. This simple setup ensures *constant error flow* in deep gradient-based error-minimizing RNNs: error signals can be backpropagated [BP1-4][BPTT1-2] through such units for millions of steps without vanishing or exploding,[VAN1] since according to the 1676 chain rule[LEI07-21b][L84] by Leibniz, the relevant multiplicative first derivatives (and their weights) are always 1.0.[VAN1]

The *invariant* residual connections transport error signals back to typically highly nonlinear *adaptive* parts of the NN where they can cause appropriate weight changes.

Note that previous self-connections with real-valued weights *other* than 1.0[MOZ] are *not* residual connections. Only 1.0 weights neutralize the vanishing/exploding gradient problem.[VAN1] However, *almost residual* connections with weights *close* to 1.0 are still acceptable in many applications. For example, a weight of 0.99 reduces an error signal backpropagated for 100 time steps (or virtual layers[BPTT1-2]) by an acceptable factor of $0.99^{100} \sim 37\%$. A weight of 0.9, however, yields only $0.9^{100} \sim 0.0027\%$.

Note that the additive weight changes of the earlier unnormalized linear Transformer (March 1991)[FWP0][ULTRA] represent a dual way of overcoming the vanishing gradient problem.[FWP]

---

## 1997 LSTM: *plain* recurrent residual connections (weight 1.0)

---

Recurrent residual connections (see above) are a defining feature of what was called Long Short-Term Memory (LSTM) in a 1995 tech report.[LSTM0,1a] The subsequent LSTM journal paper (1997)[LSTM1] has become the most cited AI paper of the 20th century.[MOST] The LSTM core units with residual connections (weight 1.0) were called *constant error carrousels* (CECs).[LSTM1] They are the very reason why LSTM can deal with huge time lags (hundreds or thousands of time steps) between inputs and target outputs. This became essential for processing speech and language.[DL4][DLH]

---

## 1999 LSTM: *gated* recurrent residual connections (gates initially open: 1.0)

---

Sometimes it is useful to let an NN modulate its residual connections through *adaptive multiplicative gates*, such that it can learn to reset itself. This was done in the *1999 LSTM variant*[LSTM2,2a] that has become known as the *vanilla LSTM*. Its so-called *forget gates* were initialised by 1.0, such that they were open, to let the LSTM start out with *plain residual connections* (weight 1.0).

Over time, the 1999 LSTM could learn when to close those gates, thus temporarily shutting down the *constant error flow*, e.g., to focus on new tasks. This reintroduces the vanishing gradient problem,[VAN1] but in a controlled way. This work was conducted by Schmidhuber's PhD student Felix Gers and his postdoc Fred Cummins.

## 2005: unfolding LSTM—from *recurrent* to *feedforward* residual NNs

The *backpropagation through time (BPTT)* algorithm[BPTT1-2][BP1-4] unfolds the sequence-processing LSTM such that it becomes a deep *feedforward* NN (FNN) with a *virtual layer* for every time step of the observed input sequence. Until 2004, the gradients of LSTM's *constant error carrousels* (CECs) were often computed by a forward method called RTRL, instead of the more storage-consuming BPTT[BPTT2][RTRL24] (back then, computational hardware was 10,000 times more expensive and much more limited than today).

In 2005, however, Schmidhuber's PhD student Alex Graves started focusing on BPTT.[LSTM3] Here the *recurrent residual connections* in the CECs become *feedforward residual connections* (weight 1.0) in a *deep residual FNN,* typically many times deeper than the unfolded FNNs of previous gradient-based RNNs, thus making LSTM many times deeper than previous RNNs. That's why LSTM can deal with much longer time lags (hundreds or thousands of time steps) between relevant observations.[DL4]

In the unfolded RNN, the resulting FNN weights are *shared across time*, but this makes no difference whatsoever for the *residual* connections: the weights of *all* residual connections in *all* RNNs and FNNs are tied to 1.0 anyway. Otherwise they wouldn't be residual connections. And whether or not the weights are shared between layers, gradients must still be propagated through many layers; hence the core role of residual connections is *identical* in both unfolded residual *RNNs* and residual *FNNs* (see below).

In RNNs, each time step/virtual layer allows for a new input and a new error signal/loss, unlike in standard FNNs. This is not an issue here. For example, in some of the original experiments designed to show LSTM's superiority (1995-1997),[LSTM0-1] there is a sequence classification loss only at the very end of each input sequence, and the error really has to travel all the way back to the first input, which is the one that makes all the difference. Again, the residual parts of the unfolded residual RNN and residual FNNs (see below) are essentially the same.

## May 2015: deep Highway Net—gated *feedforward* residual connections

While supervised LSTM RNNs had become very deep in the 1990s through residual connections, backpropagation-based FNNs had remained rather shallow until 2014: they had at most 20-30 layers or so, despite massive help through fast GPU-based hardware.[MLP1-3][DAN,DAN1][GPUCNN1-9]

Since depth is essential for deep learning, the principles of deep LSTM RNNs were transferred to deep FNNs. In May 2015, the resulting Highway Networks[HW1][HW1a] (later called *"gated ResNets"*) were the first working really deep gradient-based FNNs with hundreds of layers, over ten times deeper than previous FNNs. They worked because they adapted the 1999 LSTM principle of *gated residual connections*[LSTM2a][LSTM2] from RNNs to FNNs. This work was conducted by Schmidhuber's PhD students Rupesh Kumar Srivastava and Klaus Greff.

Let *g, t, h,* denote non-linear differentiable functions of real values. Each non-input layer of a Highway NN computes $g(x)x + t(x)h(x)$, where x is the data from the previous layer.

The *crucial residual part* is the *g(x)x* part: the gates *g(x) are typically initialised to 1.0* (like the forget gates of the 1999 LSTM[LSTM2a][LSTM2] above), to obtain *plain* residual connections (weight 1.0) which allow for very deep error propagation like in LSTM's unfolded CECs—this is what makes Highway NNs so deep.

So the initialised Highway NN starts out with very deep error propagation paths (like the later ResNet—see below). However, depending on the application, it can learn to temporarily remove the residual property of some of its residual connections in a context-dependent way, provided this improves performance. (To reduce the number of learnable parameters, the oldest Highway Net paper[HW1] actually focused on the special case *g(x) = 1 - t(x)* where *t(x)* was initialized close to 0 such that during the forward pass each layer essentially just copied its input, resulting in backpropagated error derivatives very close to 1.0—a common practice in today's deep NNs.)

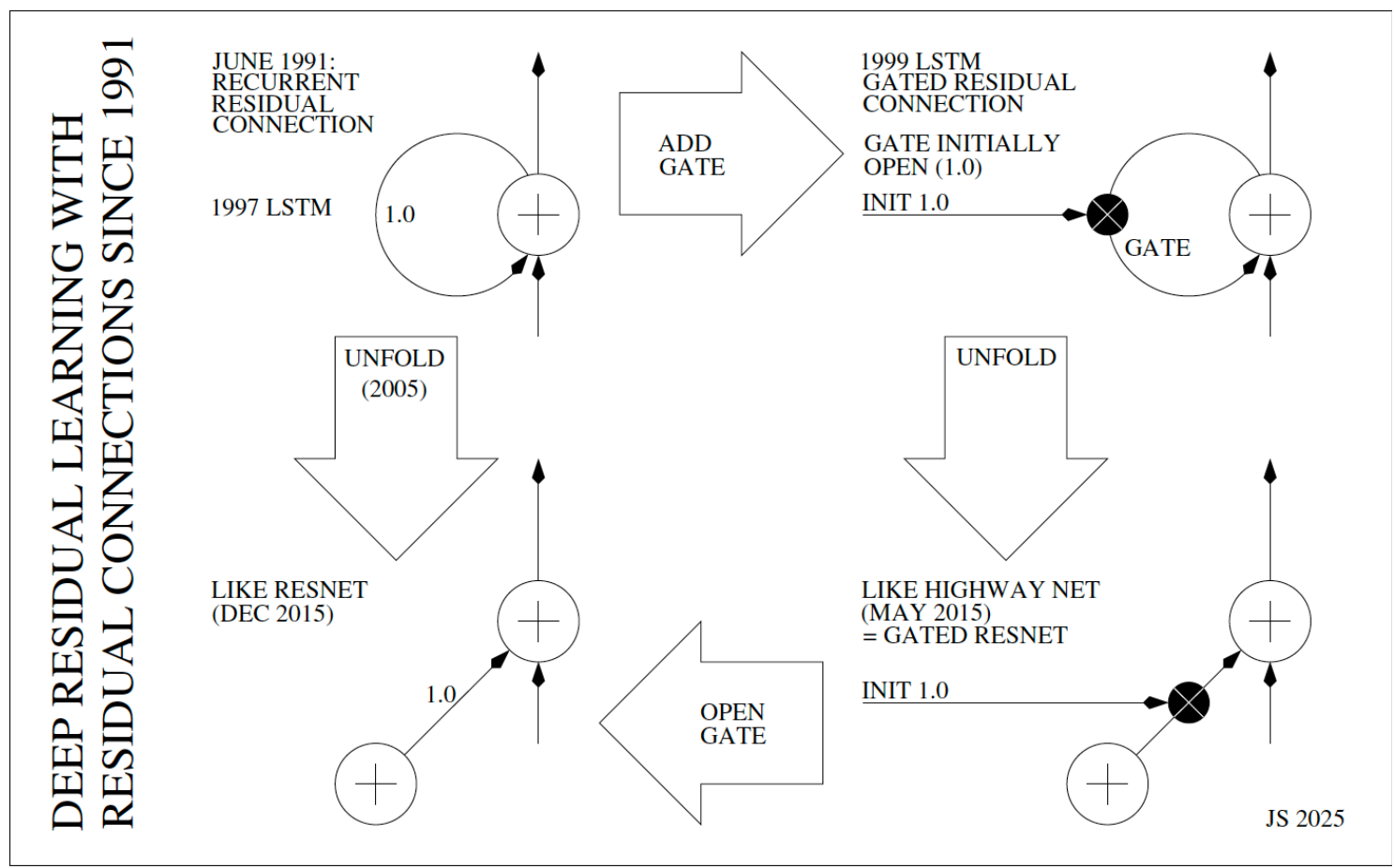

## Dec 2015: ResNet—like open-gated Highway Net (or unfolded 1997 LSTM)

Setting the Highway NN gates of May 2015[HW1][HW1a] to 1.0 at all times (not just in the initial phase of training) effectively gives us the so-called plain *Residual Net or ResNet* published 7 months later.[HW2] This *open-gated variant of the Highway Net*[HW1] is essentially a *feedforward variant of the 1997 LSTM*,[VAN1][LSTM1] while the earlier Highway Net is essentially a *gated ResNet* and a *feedforward variant of the 1999 LSTM*.[LSTM2a][LSTM2] (The term "residual" was apparently adopted from signal processing and control theory.)

Recall that the gates of the residual connections in Highway Nets are typically initialised to be open anyway, like in the 1999 LSTM. The network's training process can then decide to keep the gates open, or selectively close them if this improves performance. That is, *the residual part of the Highway Net is initialized to be like the residual part of the later ResNet*. That's what makes it so deep.

ResNets made a splash when they won the ImageNet 2015 competition.[IM15] Highway Nets perform roughly as well as ResNets on ImageNet.[HW3]

The residual parts of a ResNet look essentially like those of an unfolded 1997 LSTM (or of an initialised, open-gated 1999 LSTM).

The ResNet paper[HW2] calls the Highway Net[HW1] *"concurrent,"* but it wasn't: the ResNet was published 7 months later. The ResNet paper mentions the problem of vanishing/exploding gradients, but fails to mention that Sepp Hochreiter first identified it in 1991 and derived the solution: *residual connections*.[VAN1] The ResNet paper cites the earlier Highway Net in a way that does *not* make clear that ResNets are essentially open-gated Highway Nets, and that Highway Nets are gated ResNets, and that the gates of residual connections in Highway Nets are initially open anyway, such that Highway Nets start out with *standard* residual connections like ResNets. A follow-up paper by the ResNet authors suffered from design flaws leading to incorrect conclusions about *gated* residual connections.[HW25b]

Note again that a **residual connection** is **not** just an arbitrary **shortcut connection** or **skip connection** (e.g., 1988)[LA88][SEG1-3] from one layer to another! No, its weight must be 1.0, like in the 1997 LSTM, or in the 1999 initialized LSTM, or the initialized Highway Net, or the ResNet. If the weight had some other arbitrary real value far from 1.0, then the vanishing/exploding gradient problem[VAN1] would raise its ugly head, unless it was under control by an *initially open gate* that learns *when* to keep or temporarily remove the connection's residual property, like in the 1999 initialized LSTM, or the initialized Highway Net.

Highway NNs showed how very deep FNNs can be trained by gradient descent. This is now also relevant for Transformers[ULTRA][TR1] and other NNs. In 2021, the US Patent Office granted a patent for Highway Nets (= generalized ResNets) to our AI company NNAISENSE.

As I have often pointed out: deep learning is all about NN depth.[DL1][DLH][MIR] LSTMs brought essentially *unlimited* depth to supervised RNNs; Highway NNs brought it to FNNs. Remarkably, LSTM has become the most cited NN of the 20th century; the open-gated Highway Net variant called ResNet the most cited NN of the 21st.[MOST] The basic LSTM principle of *constant error flow through residual connections* is central not only to deep RNNs but also to deep FNNs. And it all dates back to 1991.[MIR]

---

# Acknowledgments

---

Thank you to Lucas Beyer, Francois Chollet, Kazuki Irie, Rupesh Kumar Srivastava, Sepp Hochreiter, Felix Gers, and others, for their helpful feedback. (Let me know under *juergen@idsia.ch* if you can spot any remaining error.) The contents of this

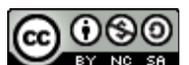

# References

[BP1] S. Linnainmaa. The representation of the cumulative rounding error of an algorithm as a Taylor expansion of the local rounding errors. Master's Thesis (in Finnish), Univ. Helsinki, 1970. *See chapters 6-7 and FORTRAN code on pages 58-60.* PDF. See also BIT 16, 146-160, 1976. Link. *The first publication on "modern" backpropagation, also known as the reverse mode of automatic differentiation.*

[BP2] P. J. Werbos. Applications of advances in nonlinear sensitivity analysis. In R. Drenick, F. Kozin, (eds): System Modeling and Optimization: Proc. IFIP, Springer, 1982. PDF. *First application of backpropagation[BP1] to NNs (concretizing thoughts in Werbos' 1974 thesis).*

[BP4] J. Schmidhuber (AI Blog, 2014; updated 2025). Who invented backpropagation? See also LinkedIn post (2025).

[BP5] A. Griewank (2012). Who invented the reverse mode of differentiation? Documenta Mathematica, Extra Volume ISMP (2012): 389-400.

[BPA] H. J. Kelley. Gradient Theory of Optimal Flight Paths. ARS Journal, Vol. 30, No. 10, pp. 947-954, 1960. *Precursor of modern backpropagation.[BP1-4]*

[BPB] A. E. Bryson. A gradient method for optimizing multi-stage allocation processes. Proc. Harvard Univ. Symposium on digital computers and their applications, 1961.

[BPC] S. E. Dreyfus. The numerical solution of variational problems. Journal of Mathematical Analysis and Applications, 5(1): 30-45, 1962.

[BPTT1] P. J. Werbos. Backpropagation through time: what it does and how to do it. Proceedings of the IEEE 78.10, 1550-1560, 1990.

[BPTT2] R. J. Williams and D. Zipser. Gradient-based learning algorithms for recurrent networks. In: Backpropagation: Theory, architectures, and applications, p 433, 1995.

[CN69] K. Fukushima (1969). Visual feature extraction by a multilayered network of analog threshold elements. IEEE Transactions on Systems Science and Cybernetics. 5 (4): 322-333. doi:10.1109/TSSC.1969.300225. *This work introduced rectified linear units or ReLUs, now widely used in CNNs and other neural nets.*

[CN79] K. Fukushima (1979). Neural network model for a mechanism of pattern recognition unaffected by shift in position—Neocognitron. Trans. IECE, vol. J62-A, no. 10, pp. 658-665, 1979. *The first deep convolutional neural network architecture, with alternating convolutional layers and downsampling layers. In Japanese. English version: [CN80]. More in Scholarpedia.*

[CN80] K. Fukushima: Neocognitron: a self-organizing neural network model for a mechanism of pattern recognition unaffected by shift in position. Biological Cybernetics, vol. 36, no. 4, pp. 193-202 (April 1980). Link.

[CN87] A. Waibel. Phoneme Recognition Using Time-Delay Neural Networks. Meeting of IEICE, Tokyo, Japan, 1987. *Application of backpropagation[BP1][BP2] and weight sharing to a 1-dimensional convolutional architecture.*

[CN87b] T. Homma, L. Atlas; R. Marks II (1987). An Artificial Neural Network for Spatio-Temporal Bipolar Patterns: Application to Phoneme Classification. Advances in Neural Information Processing Systems (N(eur)IPS), 1:31-40.

[CN88] W. Zhang, J. Tanida, K. Itoh, Y. Ichioka. Shift-invariant pattern recognition neural network and its optical architecture. Proc. Annual Conference of the Japan Society of Applied Physics, 1988. PDF. *First "modern"*

*backpropagation-trained 2-dimensional CNN, applied to character recognition.*

[CN89] W. Zhang, J. Tanida, K. Itoh, Y. Ichioka (received 13 April 1989). Parallel distributed processing model with local space-invariant interconnections and its optical architecture. Applied Optics / Vol. 29, No. 32, 1990. PDF. *First journal submission on a "modern" backpropagation-trained 2-dimensional CNN (applied to character recognition).*

[CN89b] Y. LeCun, B. Boser, J. S. Denker, D. Henderson, R. E. Howard, W. Hubbard, L. D. Jackel (received July 1989). Backpropagation Applied to Handwritten Zip Code Recognition, Neural Computation, 1(4):541-551, 1989. *Second journal submission on a "modern" backpropagation-trained 2-dimensional CNN (applied to character recognition). Compare [CN88][CN89].*

[CN89c] A. Waibel, T. Hanazawa, G. Hinton, K. Shikano and K. J. Lang. Phoneme recognition using time-delay neural networks. IEEE Transactions on Acoustics, Speech, and Signal Processing, vol. 37, no. 3, pp. 328-339, March 1989. *Based on [CN87] (1-dimensional convolutions).*

[CN89d] J. Hampshire, A. Waibel (1989). Connectionist architectures for multi-speaker phoneme recognition. In Advances in Neural Information Processing Systems, N(eur)IPS'2. *Conference publication on 2D-TDNNs or 2D-CNNs for speech recognition.*

[CN25] J. Schmidhuber (AI Blog, 2025). Who invented convolutional neural networks? See popular tweet.

[DAN] J. Schmidhuber (AI Blog, 2021). 10-year anniversary. In 2011, DanNet triggered the deep convolutional neural network (CNN) revolution. *Named after Schmidhuber's outstanding postdoc Dan Ciresan, it was the first deep and fast CNN to win international computer vision contests, and had a temporary monopoly on winning them, driven by a very fast implementation based on graphics processing units (GPUs). 1st superhuman result in 2011.[DAN1'] Now everybody is using this approach.*

[DAN1] J. Schmidhuber (AI Blog, 2011; updated 2021 for 10th birthday of DanNet): First superhuman visual pattern recognition. *At the IJCNN 2011 computer vision competition in Silicon Valley, the artificial neural network called DanNet performed twice better than humans, three times better than the closest artificial competitor (from LeCun's team), and six times better than the best non-neural method.*

[DEC] J. Schmidhuber (AI Blog, 02/20/2020, updated 2025). The 2010s: Our Decade of Deep Learning / Outlook on the 2020s. *The recent decade's most important developments and industrial applications based on our AI, with an outlook on the 2020s, also addressing privacy and data markets.*

[DL1] J. Schmidhuber, 2015. Deep learning in neural networks: An overview. Neural Networks, 61, 85-117. More. *Got the first Best Paper Award ever issued by the journal Neural Networks, founded in 1988.*

[DL4] J. Schmidhuber (AI Blog, 2017). Our impact on the world's most valuable public companies: Apple, Google, Microsoft, Facebook, Amazon... *By 2015-17, neural nets developed in Schmidhuber's labs were on over 3 billion devices such as smartphones, and used many billions of times per day, consuming a significant fraction of the world's compute. Examples: greatly improved (CTC-based) speech recognition on all Android phones, greatly improved machine translation through Google Translate and Facebook (over 4 billion LSTM-based translations per day), Apple's Siri and Quicktype on all iPhones, the answers of Amazon's Alexa, etc. Google's 2019 on-device speech recognition (on the phone, not the server) is still based on LSTM.*

[DLH] J. Schmidhuber (AI Blog, 2022). Annotated History of Modern AI and Deep Learning. Technical Report IDSIA-22-22, IDSIA, Lugano, Switzerland, 2022. Preprint arXiv:2212.11279. Tweet of 2022.

[DLP] J. Schmidhuber (AI Blog, 2023). How 3 Turing awardees republished key methods and ideas whose creators they failed to credit. Technical Report IDSIA-23-23, Swiss AI Lab IDSIA, 14 Dec 2023. Tweet of 2023.

[FWP] J. Schmidhuber (AI Blog, 26 March 2021, updated 2023, 2025). 26 March 1991: Neural nets learn to program neural nets with fast weights—like Transformer variants. 2021: New stuff! See tweet of 2022.

[FWP0] J. Schmidhuber. Learning to control fast-weight memories: An alternative to recurrent nets. Technical Report FKI-147-91, Institut für Informatik, Technische Universität München, 26 March 1991. PDF. *First paper on neural fast weight programmers that separate storage and control: a slow net learns by gradient descent to*

*compute weight changes of a fast net. The outer product-based version (Eq. 5) is now known as the unnormalized linear Transformer or the "Transformer with linearized self-attention."[ULTRA][FWP]*

[FWP1] J. Schmidhuber. Learning to control fast-weight memories: An alternative to recurrent nets. Neural Computation, 4(1):131-139, 1992. Based on [FWP0]. PDF. HTML. Pictures (German). See tweet of 2022 for 30-year anniversary.

[FWP2] J. Schmidhuber. Reducing the ratio between learning complexity and number of time-varying variables in fully recurrent nets. In Proceedings of the International Conference on Artificial Neural Networks, Amsterdam, pages 460-463. Springer, 1993. PDF. *A recurrent extension of the unnormalized linear Transformer,[ULTRA] introducing the terminology of learning "internal spotlights of attention." First recurrent NN-based fast weight programmer using outer products to program weight matrices.*

[FWP3] I. Schlag, J. Schmidhuber. Gated Fast Weights for On-The-Fly Neural Program Generation. Workshop on Meta-Learning, @N(eur)IPS 2017, Long Beach, CA, USA.

[FWP3a] I. Schlag, J. Schmidhuber. Learning to Reason with Third Order Tensor Products. Advances in Neural Information Processing Systems N(eur)IPS), Montreal, 2018. Preprint: arXiv:1811.12143. PDF.

[FWP6] I. Schlag, K. Irie, J. Schmidhuber. Linear Transformers Are Secretly Fast Weight Programmers. ICML 2021. Preprint: arXiv:2102.11174.

[FWP7] K. Irie, I. Schlag, R. Csordas, J. Schmidhuber. Going Beyond Linear Transformers with Recurrent Fast Weight Programmers. NeurIPS 2021. Preprint: arXiv:2106.06295 (June 2021).

[FWP8] K. Irie, F. Faccio, J. Schmidhuber. Neural Differential Equations for Learning to Program Neural Nets Through Continuous Learning Rules. NeurIPS 2022.

[FWP9] K. Irie, J. Schmidhuber. Images as Weight Matrices: Sequential Image Generation Through Synaptic Learning Rules. ICLR 2023.

[HW] J. Schmidhuber (AI Blog, 2015, updated 2025 for 10-year anniversary). Overview of Highway Networks: First working really deep feedforward neural networks with over 100 layers.

[HW1] R. K. Srivastava, K. Greff, J. Schmidhuber. Highway networks. Preprints arXiv:1505.00387 (May 2015) and arXiv:1507.06228 (Training Very Deep Networks; July 2015). Also at NeurIPS 2015. *The first working very deep gradient-based feedforward neural nets (FNNs) with hundreds of layers (ten times deeper than previous gradient-based FNNs). Let g, t, h, denote non-linear differentiable functions. Each non-input layer of a Highway Net computes g(x)x + t(x)h(x), where x is the data from the previous layer. The gates g(x) are typically initialised to 1.0, to obtain plain residual connections (weight 1.0). This allows for very deep error propagation, which makes Highway NNs so deep. The later Resnet (Dec 2015)[HW2] adopted this principle. It is like a Highway net variant whose gates are always open: g(x)=t(x)=const=1. That is, Highway Nets are gated ResNets: set the gates to 1.0→ResNet. Highway Nets perform roughly as well as ResNets[HW2] on ImageNet.[HW3] Variants of Highway gates are also used for certain algorithmic tasks, where plain residual layers do not work as well.[NDR] See also: who invented deep residual learning?[HW2,HW25b] More.*

[HW1a] R. K. Srivastava, K. Greff, J. Schmidhuber. Highway networks. Presentation at the Deep Learning Workshop, ICML'15, July 10-11, 2015. Link.

[HW2] He, K., Zhang, X., Ren, S., Sun, J. Deep residual learning for image recognition. Preprint arXiv:1512.03385 (Dec 2015). *Residual nets are essentially open-gated variants of the earlier very deep Highway Nets (May 2015) [HW1]. In turn, Highway Nets are generalized (gated) ResNets: set the gates to 1.0→ResNet. In fact, the gates of the residual connections in Highway Nets are typically initialised to be open (1.0) anyway, to obtain plain residual connections (weight 1.0) that permit very deep error propagation. See also: who invented deep residual learning?[HW25b,HW25b] More.*

[HW3] K. Greff, R. K. Srivastava, J. Schmidhuber. Highway and Residual Networks learn Unrolled Iterative Estimation. Preprint arxiv:1612.07771 (2016). Also at ICLR 2017.

[HW25] J. Schmidhuber (AI Blog, 2025). Who Invented Deep Residual Learning? Technical Report IDSIA-09-25, IDSIA, 2025.

[HW25b] R. K. Srivastava (January 2025). Weighted Skip Connections are Not Harmful for Deep Nets. *Shows that a follow-up paper by the authors of [HW2] suffered from design flaws leading to incorrect conclusions about gated residual connections.*

[IM15] ImageNet Large Scale Visual Recognition Challenge 2015 (ILSVRC2015): Results

[LA88] K. Lang, M. Witbrock (1988). Learning to tell two spirals apart. Proce. 1988 Connectionist Models Summer School, p. 52–59.

[L84] G. Leibniz (1684). Nova Methodus pro Maximis et Minimis. *First publication of "modern" infinitesimal calculus.*

[LEI07] J. M. Child (translator), G. W. Leibniz (Author). The Early Mathematical Manuscripts of Leibniz. Merchant Books, 2007. *See p. 126: the chain rule appeared in a 1676 memoir by Leibniz.*

[LEI10] O. H. Rodriguez, J. M. Lopez Fernandez (2010). A semiotic reflection on the didactics of the Chain rule. The Mathematics Enthusiast: Vol. 7 : No. 2 , Article 10. DOI: https://doi.org/10.54870/1551-3440.1191.

[LEI21] J. Schmidhuber (AI Blog, 2021). 375th birthday of Leibniz, founder of computer science.

[LEI21a] J. Schmidhuber (2021). Der erste Informatiker. Wie Gottfried Wilhelm Leibniz den Computer erdachte. (The first computer scientist. How Gottfried Wilhelm Leibniz conceived the computer.) Frankfurter Allgemeine Zeitung (FAZ), 17/5/2021. FAZ online: 19/5/2021.

[LEI21b] J. Schmidhuber (AI Blog, 2021). 375. Geburtstag des Herrn Leibniz, dem Vater der Informatik.

[LSTM0] S. Hochreiter and J. Schmidhuber. Long Short-Term Memory. TR FKI-207-95, TUM, August 1995. PDF.

[LSTM1a] S. Hochreiter and J. Schmidhuber. LSTM can solve hard long time lag problems. Proceedings of the 9th International Conference on Neural Information Processing Systems (NIPS'96). Cambridge, MA, USA, MIT Press, p. 473–479.

[LSTM1] S. Hochreiter, J. Schmidhuber. Long Short-Term Memory. Neural Computation, 9(8):1735-1780, 1997. PDF. Based on [LSTM0]. More.

[LSTM2a] F. A. Gers, J. Schmidhuber, F. Cummins. Learning to Forget: Continual Prediction with LSTM. In Proc. Int. Conf. on Artificial Neural Networks (ICANN'99), Edinburgh, Scotland, p. 850-855, IEE, London, 1999. *The "vanilla LSTM architecture" with forget gates that everybody is using today, e.g., in Google's Tensorflow.*

[LSTM2] F. A. Gers, J. Schmidhuber, F. Cummins. Learning to Forget: Continual Prediction with LSTM. Neural Computation, 12(10):2451-2471, 2000. PDF. *[The "vanilla LSTM architecture" that everybody is using today, e.g., in Google's Tensorflow.]*

[LSTM3] A. Graves, J. Schmidhuber. Framewise phoneme classification with bidirectional LSTM and other neural network architectures. Neural Networks, 18:5-6, pp. 602-610, 2005. PDF.

[LSTM5] A. Graves, M. Liwicki, S. Fernandez, R. Bertolami, H. Bunke, J. Schmidhuber. A Novel Connectionist System for Improved Unconstrained Handwriting Recognition. IEEE Transactions on Pattern Analysis and Machine Intelligence, vol. 31, no. 5, 2009. PDF.

[GPUCNN] K. Chellapilla, S. Puri, P. Simard. High performance convolutional neural networks for document processing. International Workshop on Frontiers in Handwriting Recognition, 2006. *Speeding up shallow CNNs on GPU by a factor of 4.*

[GPUCNN1] D. C. Ciresan, U. Meier, J. Masci, L. M. Gambardella, J. Schmidhuber. Flexible, High Performance Convolutional Neural Networks for Image Classification. *International Joint Conference on Artificial Intelligence*

*(IJCAI-2011, Barcelona)*, 2011. PDF. ArXiv preprint. *Speeding up deep CNNs on GPU by a factor of 60. Used to win four important computer vision competitions 2011-2012 before others won any with similar approaches.*

[GPUCNN2] D. C. Ciresan, U. Meier, J. Masci, J. Schmidhuber. A Committee of Neural Networks for Traffic Sign Classification. *International Joint Conference on Neural Networks (IJCNN-2011, San Francisco)*, 2011. PDF. HTML overview. *First superhuman performance in a computer vision contest, with half the error rate of humans, and one third the error rate of the closest competitor.*[DAN1] *This led to massive interest from industry.*

[GPUCNN3] D. C. Ciresan, U. Meier, J. Schmidhuber. Multi-column Deep Neural Networks for Image Classification. Proc. *IEEE Conf. on Computer Vision and Pattern Recognition CVPR 2012*, p 3642-3649, July 2012. PDF. Longer TR of Feb 2012: arXiv:1202.2745v1 [cs.CV]. More.

[GPUCNN4] A. Krizhevsky, I. Sutskever, G. E. Hinton. ImageNet Classification with Deep Convolutional Neural Networks. NIPS 25, MIT Press, Dec 2012. PDF. *This paper describes AlexNet, which is similar to the earlier DanNet,*[DAN,DAN1][R6] *the first pure deep CNN to win computer vision contests in 2011*[GPUCNN3-5,5] *(AlexNet and VGG Net*[GPUCNN9]*) followed in 2012-2014). [GPUCNN4] emphasizes benefits of Fukushima's ReLUs (1969)*[CN69] *and dropout (a variant of Hanson 1990 stochastic delta rule)*[Drop1-4] *but neither cites the original work*[CN69][Drop1] *nor the basic CNN architecture (Fukushima, 1979).*[CN79][CN25]

[GPUCNN5] J. Schmidhuber (AI Blog, 2017; updated 2021 for 10th birthday of DanNet): History of computer vision contests won by deep CNNs since 2011. DanNet was the first CNN to win one, and won 4 of them in a row before the similar AlexNet/VGG Net and the Resnet (a Highway Net with open gates) joined the party. Today, deep CNNs are standard in computer vision.

[GPUCNN6] J. Schmidhuber, D. Ciresan, U. Meier, J. Masci, A. Graves. On Fast Deep Nets for AGI Vision. In Proc. Fourth Conference on Artificial General Intelligence (AGI-11), Google, Mountain View, California, 2011. PDF.

[GPUCNN7] D. C. Ciresan, A. Giusti, L. M. Gambardella, J. Schmidhuber. Mitosis Detection in Breast Cancer Histology Images using Deep Neural Networks. MICCAI 2013. PDF.

[GPUCNN8] J. Schmidhuber (AI Blog, 2017; updated 2021 for 10th birthday of DanNet). First deep learner to win a contest on object detection in large images— first deep learner to win a medical imaging contest (2012). Link. *How the Swiss AI Lab IDSIA used GPU-based CNNs to win the ICPR 2012 Contest on Mitosis Detection and the MICCAI 2013 Grand Challenge.*

[GPUCNN9] K. Simonyan, A. Zisserman. Very deep convolutional networks for large-scale image recognition. Preprint arXiv:1409.1556 (2014).

[MIR] J. Schmidhuber (Oct 2019, updated 2021, 2022, 2025). Deep Learning: Our Miraculous Year 1990-1991. Preprint arXiv:2005.05744. *The Deep Learning Artificial Neural Networks (NNs) of our team have revolutionised Machine Learning & AI. Many of the basic ideas behind this revolution were published within the 12 months of our "Annus Mirabilis" 1990-1991 at our lab in TU Munich. Back then, few people were interested, but a quarter century later, NNs based on our "Miraculous Year"* were on over 3 billion devices, and used many billions of times per day, consuming a significant fraction of the world's compute. *In particular, in 1990-91, we laid foundations of Generative AI, publishing principles of (1)* Generative Adversarial Networks *for* Artificial Curiosity and Creativity *(now used for deepfakes), (2)* Transformers *(the T in ChatGPT—see the* 1991 Unnormalized Linear Transformer*), (3)* Pre-training *for deep NNs (see the P in ChatGPT), (4)* NN distillation *(key for* DeepSeek*), and (5) recurrent* World Models *for* Reinforcement Learning and Planning *in partially observable environments. The year 1991 also marks the emergence of the defining features of (6)* LSTM, *the most cited AI paper of the 20th century (based on constant error flow through residual NN connections), and (7) ResNet, the most cited AI paper of the 21st century, based on our LSTM-inspired* Highway Net *that was 10 times deeper than previous feedforward NNs.*

[MLP1] D. C. Ciresan, U. Meier, L. M. Gambardella, J. Schmidhuber. Deep Big Simple Neural Nets For Handwritten Digit Recognition. Neural Computation 22(12): 3207-3220, 2010. ArXiv Preprint. *Showed that plain backprop for deep standard NNs is sufficient to break benchmark records, without any unsupervised pre-training.*

[MLP3] J. Schmidhuber (AI Blog, 2025). 2010: Breakthrough of end-to-end deep learning (no layer-by-layer training, no unsupervised pre-training). The rest is history. *By 2010, when compute was 1000 times more expensive than in 2025, both our feedforward NNs*[MLP1] *and our earlier recurrent NNs were able to beat all*

*competing algorithms on important problems of that time. This deep learning revolution quickly spread from Europe to North America and Asia.*

[MOST] J. Schmidhuber (AI Blog, 2021, updated 2025). The most cited neural networks all build on work done in my labs: *1. Long Short-Term Memory (LSTM), the most cited AI of the 20th century. 2. ResNet (open-gated Highway Net), the most cited AI of the 21st century. 3. AlexNet & VGG Net (the similar but earlier DanNet of 2011 won 4 image recognition challenges before them). 4. GAN (an instance of Adversarial Artificial Curiosity of 1990). 5. Transformer variants—see the 1991 unnormalised linear Transformer (ULTRA). Foundations of Generative AI were published in 1991: the principles of GANs (now used for deepfakes), Transformers (the T in ChatGPT), Pre-training for deep NNs (the P in ChatGPT), NN distillation, and the famous DeepSeek—see the tweet.*

[MOST25] H. Pearson, H. Ledford, M. Hutson, R. Van Norden. Exclusive: the most-cited papers of the twenty-first century. Nature, 15 April 2025.

[MOST25b] R. Van Norden. Science's golden oldies: the decades-old research papers still heavily cited today. Nature, 15 April 2025.

[MOZ] M. Mozer. A Focused Backpropagation Algorithm for Temporal Pattern Recognition. Complex Systems, 1989.

[NDR] R. Csordas, K. Irie, J. Schmidhuber. The Neural Data Router: Adaptive Control Flow in Transformers Improves Systematic Generalization. Proc. ICLR 2022. Preprint arXiv/2110.07732.

[NOB] J. Schmidhuber. A Nobel Prize for Plagiarism. Technical Report IDSIA-24-24 (7 Dec 2024, updated 31 July 2025). *Sadly, the Nobel Prize in Physics 2024 for Hopfield & Hinton is a Nobel Prize for plagiarism. They republished methodologies for artificial neural networks developed in Ukraine and Japan by Ivakhnenko and Amari in the 1960s & 1970s, as well as other techniques, without citing the original papers. Even in later surveys, they didn't credit the original inventors (thus turning what may have been unintentional plagiarism into a deliberate form). None of the important algorithms for modern Artificial Intelligence were created by Hopfield & Hinton. See also popular tweet1, tweet2, and LinkedIn post.*

[RTRL24] K. Irie, A. Gopalakrishnan, J. Schmidhuber Exploring the Promise and Limits of Real-Time Recurrent Learning. ICLR 2024.

[SEG1] J. Long, E. Shelhamer, T. Darrell. Fully Convolutional Networks for Semantic Segmentation. CVPR 2015. Preprint arXiv:1411.4038 (v1 2014).

[SEG2] B. Hariharan, P. Arbelaez, R. Girshick, J. Malik. Hypercolumns for Object Segmentation and Fine-grained Localization. CVPR 2015. Preprint arXiv:1411.5752 (v1 2014).

[SEG3] O. Ronneberger, P. Fischer , T. Brox (2015). U-Net: Convolutional Networks for Biomedical Image Segmentation. Preprint arXiv:1505.04597.

[TR1] A. Vaswani, N. Shazeer, N. Parmar, J. Uszkoreit, L. Jones, A. N. Gomez, L. Kaiser, I. Polosukhin (2017). Attention is all you need. NIPS 2017, pp. 5998-6008. *This paper introduced the name "Transformers" for a now widely used NN type. It did not cite the 1991 publication on what's now called unnormalized "linear Transformers" with "linearized self-attention."[ULTRA] Schmidhuber also introduced the now popular attention terminology in 1993.[ATT][FWP2][R4] See tweet of 2022 for 30-year anniversary.*

[TR2] J. Devlin, M. W. Chang, K. Lee, K. Toutanova (2018). Bert: Pre-training of deep bidirectional Transformers for language understanding. Preprint arXiv:1810.04805.

[TR3] K. Tran, A. Bisazza, C. Monz. The Importance of Being Recurrent for Modeling Hierarchical Structure. EMNLP 2018, p 4731-4736. ArXiv preprint 1803.03585.

[TR4] M. Hahn. Theoretical Limitations of Self-Attention in Neural Sequence Models. Transactions of the Association for Computational Linguistics, Volume 8, p.156-171, 2020.

[TR5] A. Katharopoulos, A. Vyas, N. Pappas, F. Fleuret. Transformers are RNNs: Fast autoregressive Transformers with linear attention. In Proc. Int. Conf. on Machine Learning (ICML), July 2020.

[TR5a] Z. Shen, M. Zhang, H. Zhao, S. Yi, H. Li. Efficient Attention: Attention with Linear Complexities. WACV 2021.

[TR6] K. Choromanski, V. Likhosherstov, D. Dohan, X. Song, A. Gane, T. Sarlos, P. Hawkins, J. Davis, A. Mohiuddin, L. Kaiser, et al. Rethinking attention with Performers. In Int. Conf. on Learning Representations (ICLR), 2021.

[TR6a] H. Peng, N. Pappas, D. Yogatama, R. Schwartz, N. A. Smith, L. Kong. Random Feature Attention. ICLR 2021.

[TR7] S. Bhattamishra, K. Ahuja, N. Goyal. On the Ability and Limitations of Transformers to Recognize Formal Languages. EMNLP 2020.

[TR8] W. Merrill, A. Sabharwal. The Parallelism Tradeoff: Limitations of Log-Precision Transformers. TACL 2023.

[ULTRA] References on the 1991 unnormalized linear Transformer (ULTRA): original tech report (March 1991) [FWP0]. Journal publication (1992) [FWP1]. Recurrent ULTRA extension (1993) introducing the terminology of learning "internal spotlights of attention" [FWP2]. Modern *"quadratic"* Transformer (2017: *"attention is all you need"*) scaling *quadratically* in input size [TR1]. 2020 paper [TR5] using the terminology *"linear Transformer"* for a more efficient Transformer variant that scales *linearly*, leveraging *linearized attention* [TR5a]. 2021 paper [FWP6] pointing out that ULTRA dates back to 1991 [FWP0] when compute was a million times more expensive. Overview of ULTRA and other Fast Weight Programmers (2021) [FWP]. See the T in ChatGPT! See also surveys [DLH][DLP], 2022 tweet for ULTRA's 30-year anniversary, and 2024 tweet.

[UN] J. Schmidhuber (AI Blog, 2021). 30-year anniversary. 1991: First very deep learning with unsupervised or self-supervised pre-training. *Unsupervised hierarchical predictive coding (with self-supervised target generation) finds compact internal representations of sequential data to facilitate downstream deep learning. The hierarchy can be distilled into a single deep neural network (suggesting a simple model of conscious and subconscious information processing). 1993: solving problems of depth >1000.*

[VAN1] S. Hochreiter. Untersuchungen zu dynamischen neuronalen Netzen. Diploma thesis, TUM, 1991 (advisor J. Schmidhuber). PDF. *More on the Fundamental Deep Learning Problem.*

[VAN2] Y. Bengio, P. Simard, P. Frasconi. Learning long-term dependencies with gradient descent is difficult. IEEE TNN 5(2), p 157-166, 1994. *Results are essentially identical to those of Schmidhuber's diploma student Sepp Hochreiter (1991).[VAN1] Even after a common publication,[VAN3] the first author of [VAN2] published papers that cited only their own but not the original work.[DLP]*

[VAN3] S. Hochreiter, Y. Bengio, P. Frasconi, J. Schmidhuber. Gradient flow in recurrent nets: the difficulty of learning long-term dependencies. In S. C. Kremer and J. F. Kolen, eds., A Field Guide to Dynamical Recurrent Neural Networks. IEEE press, 2001. PDF.

.